\documentclass[10pt,twocolumn,letterpaper]{article}

\usepackage{cvpr}              
\usepackage{float}
\usepackage[pagebackref,breaklinks,colorlinks,allcolors=cvprblue, colorlinks=true,linkcolor=red,citecolor=green]{hyperref}

\newcommand{\red}[1]{{\color{red}#1}}

\definecolor{cvprblue}{rgb}{0.21,0.49,0.74}
\usepackage{multirow}

\def\paperID{11647} 
\def\confName{CVPR}
\def\confYear{2025}

\title{Just Dance with $\pi$! \\ A \underline{P}oly-modal \underline{I}nductor for Weakly-supervised Video Anomaly Detection}

 \author{\parbox{16cm}{\centering
    {\large Snehashis Majhi$^{1,2{\color{red}*}}$,  Giacomo D'Amicantonio$^{5{\color{red}*}}$, Antitza Dantcheva$^{1,2}$, Quan Kong$^3$, Lorenzo Garattoni$^4$,  Gianpiero Francesca$^4$, Egor Bondarev$^5$, François Brémond$^{1,2}$}\\
    {\normalsize
    $^1$ INRIA \quad
    $^2$ Côte d'Azur University \quad
    $^3$ Woven by Toyota\quad
    $^4$ Toyota Motor Europe \quad
    $^5$ Eindhoven University of Technology}}\\
    \vspace{0.3cm}
    \small{{\color{red}* Joint first authors.}} \quad \small{{{Code: \url{https://github.com/snehashismajhi/PI-VAD}}}}
    \vspace{-0.7cm}
}

\begin{document}

\twocolumn[{
\maketitle
\begin{center}
   \captionsetup{type=figure}
   \includegraphics[width=\textwidth]{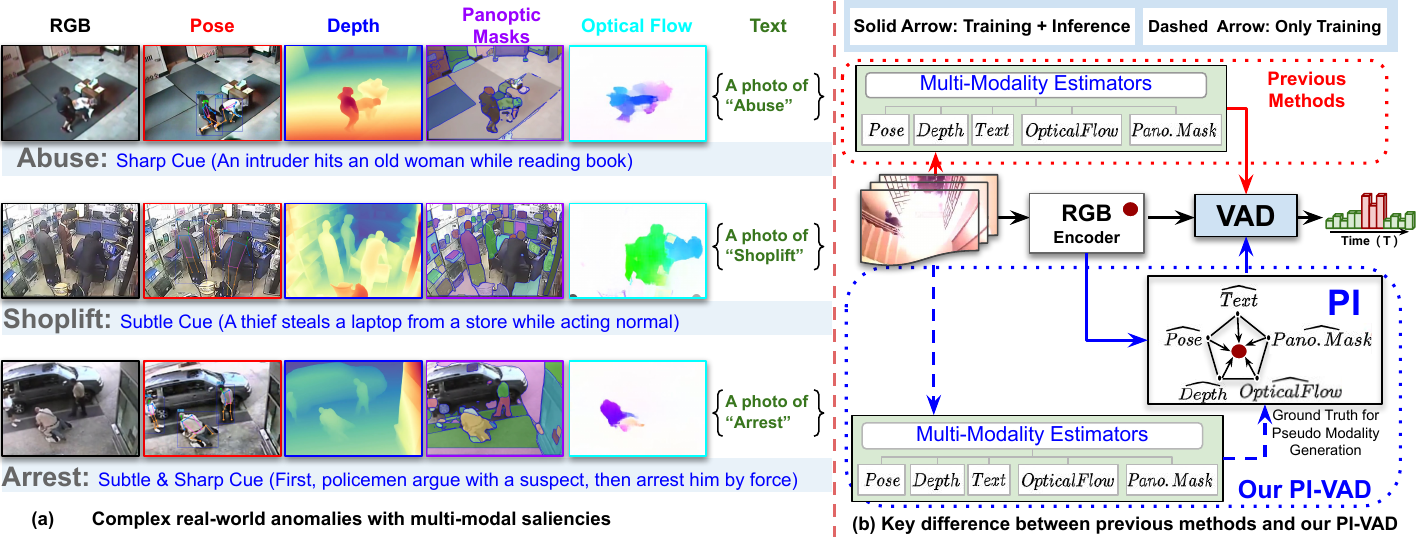}
    \vspace{-0.5cm}
   \captionof{figure}{\textbf{a):} Illustration of abnormal frames and respective multi-modal saliencies in complex real-world scenes. \textit{Optical flow} captures distinct abnormal motion in ``Abuse” and ``Arrest”, while \textit{depth} and \textit{pose} detect subtle movements that \textit{optical flow} may miss. \textit{Panoptic masks} and \textit{text} provide overall scene context. \textbf{b):} Comparison of multi-modal methods with our PI-VAD. PI-VAD requires the five modalities only during training, significantly reducing computation and enabling real-world applicability.}
\end{center}
}]

\maketitle

\begin{abstract}

\vspace{-0.4cm}

Weakly-supervised methods 
for video anomaly detection (VAD) 
are conventionally based merely on RGB spatio-temporal features, which continues to limit their reliability 
in real-world scenarios. This is due to the fact that RGB-features are not sufficiently distinctive in setting apart categories such as \textit{shoplifting} from visually similar events. Therefore, 
towards robust complex real-world VAD, 
it is essential to augment RGB spatio-temporal features by additional modalities. 
Motivated by this, we introduce 
the Poly-modal Induced framework for VAD: ``\textbf{PI-VAD}" (or $\pi$-VAD), a novel approach that augments RGB representations by five additional modalities. Specifically, the modalities include sensitivity to fine-grained motion (Pose), three dimensional scene and entity representation (Depth), surrounding objects (Panoptic masks), global motion (optical flow), as well as language cues (VLM). Each modality represents an axis of a polygon, streamlined to add salient cues to RGB. 
$\pi$-VAD includes two plug-in modules, namely Pseudo-modality Generation module and Cross Modal Induction module, which generate modality-specific prototypical representation and, thereby, induce multi-modal information into RGB cues. These modules operate by performing anomaly-aware auxiliary tasks and necessitate five modality backbones -- only during training. Notably, $\pi$-VAD achieves state-of-the-art accuracy on three prominent VAD datasets encompassing real-world scenarios, without requiring the computational overhead of five modality backbones at inference.

\end{abstract}    
\vspace{-0.5cm}
\section{Introduction}
\label{sec:intro}

Weakly supervised video anomaly detection (WSVAD) aims to predict frame-level anomaly scores using only video-level labels, avoiding the need for detailed frame-by-frame annotation. WSVAD methods~\cite{cvpr18,cvpr19,eccv2020,CVPR23LAA, iccv21_Dance} are effective for detecting large-scale scene anomalies, like explosions or road accidents, by training on both normal and anomalous videos to improve generalization in diverse real-world settings. However, they often struggle with more complex, human-centered anomalies such as ``shoplifting'', ``stealing'', and ``abuse'', where human interactions and subtle actions are involved. This limitation stems from the fact that most current methods rely on single-modality (video-only) features, which may not fully capture the complexity of these scenarios.
Towards improving WSVAD in real-world settings, we place emphasis on including additional modalities, such as \textit{pose}, \textit{depth}, \textit{panoptic masks}, \textit{optical flow}, and \textit{language semantics}, to facilitate a more nuanced scene representation. The additional information provided by these modalities describes detailed human movements, entity distances, motion dynamics, and context, rendering WSVAD more effective for detecting complex anomalies.

Despite affirmative implications of multi-modal semantics, their adaptation to WSVAD remains under-explored in current research. This is majorly due to three reasons: \textbf{(i)} \textbf{limited data with limited supervision:} while recent multi-modal foundation models like CLIP~\cite{CLIP}, IMAGEBIND~\cite{imagebind} necessitates more than 400 million images for multi-modal association, the anomaly-detection task inherently deals with sparse and limited data (e.g $810$ anomaly videos in UCF-Crime dataset~\cite{cvpr18}). Further, the absence of frame-level labels in WSVAD can lead to ambiguous multi-modal association; \textbf{(ii) disparity among modalities:} since each modality captures unique characteristics at various semantic levels (i.e. from contextual to fine-grained), there exists an underlying disparity among modalities that brings to the fore additional challenges in associating the modalities meaningfully; \textbf{(iii) increased inference overhead:} common multi-modal foundation models presume the availability of all modalities during inference as well, thereby linearly adding multiple modalities to the framework increases the inference overhead significantly, hindering real-time applicability. 
These challenges lead us to the main question: \textbf{what is the best strategy to combine multiple disparate modalities to RGB with limited data and supervision, without compromising the latency?}


%
Motivated by the above, we introduce a novel \underline{P}oly-modal \underline{I}nduced Transformer for weakly-supervised video anomaly detection, called \textbf{PI-VAD} (or \textbf{$\pi$-VAD}). Deviating from all WSVAD benchmarks, $\pi$-VAD synthesizes latent embeddings from five complementary modalities — pose, depth, panoptic segmentation, optical flow, and language semantics — to augment and enrich RGB-based analysis. $\pi$-VAD comprises two novel plugin modules that integrate seamlessly into a WSVAD framework: \textbf{(i)} the Pseudo Modality Generation (PMG) module, and \textbf{(ii)} the Cross Modal Induction (CMI) module. The PMG module generates synthetic, modality-specific prototype embeddings directly from RGB features, capturing each modality's distinctive characteristics. This approach mitigates inference latency by circumventing the need for individual modality backbones, thus preserving $\pi$-VAD operational efficiency.


The CMI module aligns uncoupled modalities within a unified, RGB-anchored embedding space through a double-alignment process. Initially, it constructs semantic associations between each modality and RGB via a contrastive alignment objective, ensuring cohesive integration of multi-modal embeddings. CMI leverages a pre-trained VAD model to guide the aligned multi-modal representations towards a unified task-aware and aligned representation, ensuring that the learned alignments are contextually relevant to anomaly detection. This distillation process injects $\pi$-VAD with a nuanced, semantically grounded multi-modal representation, enabling robust anomaly detection even under limited data and supervision. Moreover, the architecture of $\pi$-VAD facilitates the scalable incorporation of additional modalities without exacerbating latency constraints. To our knowledge, $\pi$-VAD is the first framework to harness the full spectrum of multi-modal representations within WSVAD, setting a new paradigm for complex anomaly detection in video analysis.

To summarize, our contributions are three-fold.
\begin{itemize}
    \item We introduce $\pi$-VAD, a novel multi-modal method that harnesses five or more modalities to seamlessly infuse critical multi-modal cues into RGB cues, thereby enhancing the weakly-supervised video anomaly detection.
    \item We present two-plugin modules that are designed to synthesize multi-modal prototypes and learn effective associations to RGB. These plugin modules perform anomaly-aware auxiliary task to generate and bind meaningful multi-modal representations
    \item We provide an exhaustive experimental analysis to validate the robustness of $\pi$-VAD on UCF-Crime~\cite{cvpr18}, XD-Violence~\cite{eccv2020}, and MSAD~\cite{msad2024}. 
    The results suggest that $\pi$-VAD outperforms previous prominent approaches.
\end{itemize}



\section{Related Work}
\label{sec:formatting}
\textbf{Weakly supervised video anomaly detection methods}~\cite{cvpr18,cvpr19,eccv2020,icip2019,CVPR23LAA,bmvc19,avss2019,spl2020, majhi_FG21, majhi_avss21, iccv21_Dance, ICME2020} rely on training models with video-level weak annotations, which include both normal and anomalous data. The foundational work by Sultani et al.~\cite{cvpr18} introduced a deep multiple instance learning (MIL) ranking framework for video anomaly detection. 
Since then, numerous adaptations of this approach have been developed. For instance, Tian et al.~\cite{iccv21} introduced a feature magnitude learning function to better identify anomalous instances. Chen et al.~\cite{AAAI23MGFN} proposed a feature amplification mechanism with a amplitude contrast loss to enhance the discriminative capabilities of features. Lv et al.~\cite{CVPR23UMIL} introduced an Unbiased Multiple Instance Learning (UMIL) framework to create unbiased anomaly classifiers. However, these one-stage methods often concentrate on highly discriminative segments while overlooking the ambiguous and subtle ones. To address this, recent work has shifted towards pseudo-label-based, two-stage self-training methods~\cite{AAAI22MSL, CVPR23LAA} to improve the accuracy of anomaly scores. Li et al.~\cite{AAAI22MSL} introduced a multi-sequence learning technique to iteratively refine anomaly scores by progressively shortening selected sequences. However, these methods rely on single-modal video information and do not incorporate corresponding multi-modal data. Recently, cross-modal approaches have started to incorporate information from multiple modalities to improve the accuracy of discriminative features and pseudo-labels, although they primarily use text-based anomaly categories and miss the richer semantic information of anomalous events.

\textbf{Multi-modal video representation learning} leverages multiple modalities—such as RGB, depth, text, audio, and poses—to create richer representations. This approach is commonly based on two techniques: contrastive loss and knowledge distillation. Contrastive loss, as used in models like CLIP~\cite{CLIP}, creates a shared embedding space across different modalities by aligning similar features closely. Knowledge distillation, on the other hand, transfers knowledge between modalities, allowing a salient modality like text to help a modality like RGB learn more effectively. Some methods, like ViFiCLIP~\cite{rasheed2023fine} and CoCLR~\cite{Scale_CLIP}, combine contrastive learning with knowledge distillation to fine-tune alignments across modalities, while also making cross-modal learning more efficient. However, these techniques require large-scale datasets to effectively learn multi-modal representations, while video anomaly datasets are inherently sparse and small scale. To leverage cross-modal information with limited data, we generate pseudo-modalities during training and apply contrastive loss and knowledge distillation to guide the shared feature space semantically.

\begin{figure*}[ht]
    \subfloat[Overview of $\pi$-VAD]{\includegraphics[width=0.45\textwidth]{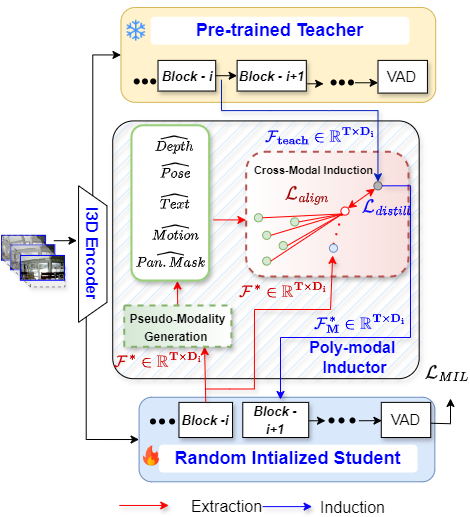}\label{fig:pivad}}\hfill
    \subfloat[Poly-modal Inductor (PI)]{\includegraphics[width=0.55\textwidth]{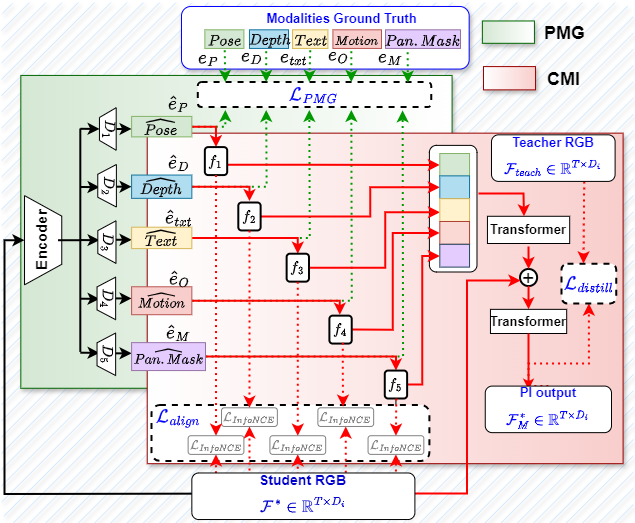}\label{fig:pi}}\hfill

    \caption{\textbf{(a) Overview of Poly-modal Induced VAD ($\pi$-VAD):} In the \textit{training phase}, $\pi$-VAD uses a teacher-student approach, where a poly-modal inductor enhances the student's RGB representation by generating and associating five distinct modalities. Note that the teacher’s weights remain fixed during training. At inference, the student and poly-modal inductor operate independently to detect video anomalies. \textbf{(b) Poly-modal Inductor (PI):} PI refines the student’s intermediate feature, $\mathcal{F}^*$, by generating pseudo-modalities through a modality generation module (PMG). These generated modalities are then combined with $\mathcal{F}^*$ to produce an enhanced feature set, $\mathcal{F}^*_M$.}
    \label{fig:inductor}
    \vspace{-0.5cm}
\end{figure*}

\section{Preliminaries: Uni-modal WSVAD Method}
In this section, we briefly describe the working principle of existing uni-modal WSVAD methods. Uni-modal WSVAD focuses solely on the RGB modality for both training and inference. 
The video \( V \) is first divided into non-overlapping snippets of 16 frames, resulting in \( T \) snippets. A pre-trained 3D convolutional network (e.g., I3D~\cite{i3d}) is then used to extract features from each snippet, forming a feature map \( \mathcal{F}_{RGB} \in \mathbb{R}^{T \times D} \), where 
\( D \) is the feature dimension. Given $\mathcal{F}_{RGB}$, the goal of the uni-modal WSVAD method is to train an RGB task encoder that can predict frame-level anomaly scores while only having access to video-level labels during training.

Deviating from standard uni-modal WSVAD, in this work we explore the multi-modal (\textit{i.e.} two or more modalities) representation learning ability in WSVAD task. We aim to answer questions such as: \textbf{how many modalities are required to represent real-world complex anomalies?} \textbf{With limited dataset and supervision, is it possible for a model to effectively learn from more than five modalities and use only RGB for inference?}While multi-modal methods of action understanding~\cite{VPNplusplus,RGB_aug_pose_flow,vpn, pivit} can be applicable to the WSVAD task, their effectiveness depends on the amount of labeled data available.
Therefore, we propose a novel multi-modal framework $\pi$-VAD that can effectively associate more than five modalities to RGB for the WSVAD task. Further, we observe that the closest prior work with our proposition is $\pi$-ViT~\cite{pivit}. Next, we proceed to outline the key differences between $\pi$-ViT and $\pi$-VAD.

    \paragraph{$\pi$-ViT Vs. $\pi$-VAD:} Both $\pi$-ViT and $\pi$-VAD pioneer the integration of RGB with additonal modalities through induction mechanisms, yet $\pi$-VAD pushes the boundaries further by incorporating over five modalities. Confronted with the complexities of handling extensive multi-modal data, $\pi$-VAD addresses challenges such as multi-modal disparity by leveraging autoencoder latent spaces to store latent prototype vectors. Distinct from $\pi$-ViT, which is optimized for fully-supervised action recognition, $\pi$-VAD is tailored for temporal video anomaly detection in scenarios constrained by limited data and supervision. To achieve robust multi-modal fusion under these constraints, $\pi$-VAD combines autoencoder-based representations with a contrastive learning objective, followed by a teacher-student training paradigm enriched by distillation loss, ensuring superior adaptability and binding across modalities.

\section{Proposed $\pi$-VAD}
In this section, we introduce our \underline{P}oly-modal \underline{I}nduced Transformer for weakly-supervised video anomaly detection, referred to as $\pi$-VAD (illustrated in Figure~\ref{fig:inductor}). $\pi$-VAD adopts a teacher-student architecture incorporating a novel poly-modal inductor. While the teacher and student share an identical functional architecture, the teacher is pre-trained on the WSVAD task and remains frozen, and the student is randomly initialized.

\subsection{Poly-modal Inductor (PI)}
The objective of the poly-modal inductor (illustrated in Figure~\ref{fig:pivad}) is to enhance the student’s RGB representation by promoting the learning of discriminative features for anomalous events within a cohesive multi-modal feature space. This is enforced by two key modules of poly-modal inductor: \textbf{(i) Pseudo Modality Generation (PMG) module} learns anomaly relevant synthetic approximation of the actual modalities component, \textbf{(ii) Cross Modal Induction (CMI) module} facilitates the semantic alignment between the multi-modal encodings from PMG and the RGB embeddings of the student while ensuring that the alignment is pertinent to WSVAD. As visible in Figure~\ref{fig:pi}, the poly-modal inductor is adaptable and can be integrated at various blocks of the teacher-student architecture; however, we deliberately position it in the initial and final blocks to capture both low and high-level multi-modal features effectively. Further,  regardless of the student’s specific block, the poly-modal inductor processes the output representation from \textit{Block-i} of the student \( \mathcal{F}^{*} \in \mathbb{R}^{T \times D_i} \) and injects the refined multi-modal feature \( \mathcal{F}^{*}_{M} \in \mathbb{R}^{T \times D_{i}} \) into \textit{Block i+1} of the student, thereby enhancing the student's ability to learn discriminative representation for anomaly detection. 

\subsection{Pseudo Modality Generation Module}
The Pseudo Modality Generation (PMG) module aims to synthetically derive embeddings for pose (\( \hat{e_{P}} \in \mathbb{R}^{T \times d_P} \)), depth (\( \hat{e_{D}} \in \mathbb{R}^{T \times d_D} \)), panoptic masks (\( \hat{e_{M}} \in \mathbb{R}^{T \times d_M} \)), optical flow (\( \hat{e_{O}} \in \mathbb{R}^{T \times d_O} \)), and text (\( \hat{e_{txt}} \in \mathbb{R}^{T \times d_{txt}} \)) directly from the student’s intermediate RGB feature representation \( \mathcal{F}^{*} \in \mathbb{R}^{T \times D_i} \). This approach fulfills two key objectives: \textbf{(i)} eliminating the reliance on multi-modal backbones (e.g., SAM, yolov7, etc.) within the poly-modal inductor during inference; \textbf{(ii)} since multi-modal embeddings can introduce redundancy, noise, or even conflicting information, this approach selectively retains only the multi-modal cues essential to the WSVAD task.

To enforce these objectives, we design PMG that can be seen as an encoder-decoder structure. As shown in Figure~\ref{fig:inductor}, PMG has one encoder and five parallel decoders $D_1$, $D_2$, $\ldots$, $D_5$. We deliberately kept one encoder to learn shared RGB features for all the modalities. The six decoders operate in a mutually exclusive manner to generate the six modalities. The encoder has a 1D-convolutional layer to project RGB embeddings to a low-dimensional latent space.

For each modality decoder, a single linear layer translates the RGB latent representation to a modality-specific RGB representation, maintaining the latent space's dimensionality. By doing this, we generate diverse views of the same RGB embeddings, enhancing the information contained in the embeddings that is relevant to a specific modality while suppressing possible noise. Subsequently, a 1D convolution layer is used as a decoder to generate the modality embeddings $\hat{e_j}$, where $j \in \{P, D, M, O, \text{txt}\}$. 

Training the PMG requires \textit{ground-truth} embeddings $e_j$, where $j \in \{P, D, M, O, \text{txt}\}$ from the corresponding modality decoders. We utilize the intermediate embeddings of YOLOV7-pose \cite{wang2023yolov7}, DepthAnythingV2 \cite{yang2024depth}, SAM \cite{kirillov2023segment}, RAFT \cite{teed2020raft} and VifiCLIP~\cite{rasheed2023fine} to represent pose, depth, panoptic mask, optical flow and text modality \textit{ground-truths}. The combined training objective for the PMG is
\begin{equation}
\mathcal{L}_{PMG} = \sum_{j=1}^{5}\frac{1}{d_j}\sum_{k=1}^{d_j}(\hat{e_{j,k}}-e_{j,k})^2, \small \text{where } j \in \{\text{P, D, M, O, txt}\}.
\end{equation}

Once PMG is trained with $\mathcal{L}_{PMG}$, it has the ability to precisely generate pseudo modalities which are subsequently used in PI to augment the RGB representation of the student.

\subsection{Cross Modal Induction Module}
In this stage, Cross Modal Induction (CMI) combines the generated pseudo-modalities \( \hat{e_j} \) 
with the RGB embeddings \( \mathcal{F}^{*} \), aiming to create a shared representation space that promotes all relevant features for the task. It aligns the pseudo-modalities from the PMG with the RGB embeddings, which contain critical visual information from the current video snippet \( T_i \). Our aim is to ensure that the most relevant modalities for \( T_i \) converge with the RGB embeddings in a cohesive representation space, thereby strengthening multi-modal associations. By aligning these diverse modalities, the PMG produces modality embeddings informed by the RGB data, enhancing relevant information and filtering out irrelevant details. This joint representation is essential for the WSVAD task, as it improves the model's ability to utilize multi-modal insights effectively.


To achieve this, we learn a shared latent space between each modality and the RGB embeddings by applying a snippet-level, bi-directional InfoNCE contrastive loss~\cite{oord2018representation}. This loss is applied between each pseudo-modality embedding \( \hat{e_j} \) (where \( j \in \{P, D, M, O, \text{txt}\} \)) and the RGB embedding \( \mathcal{F}^{*} \). The bi-directional approach provides a more balanced measure of similarity between positive and negative pairs. Since the contrastive loss is applied at the snippet level, we treat representations from the same snippet index \( T_i \) (i.e., \( \mathcal{F}^{*}(T_i) \) and \( \hat{e_j}(T_i) \)) as positive pairs, and representations from different snippets as negative pairs. This encourages similarity in positive pairs and discourages similarity in negative pairs. The similarity between embeddings is computed as: $sim(\mathcal{F}^{*} (T_i), \hat{e_j} (T_i)) = \frac{\mathcal{F}^{*} (T_i) \cdot \hat{e_j} (T_i)}{\|\mathcal{F}^{*} (T_i)\| \|\hat{e_j} (T_i)\|}$ and the contrastive alignment loss is defined as


\vspace{-0.3cm}
\begin{equation}
\small
\mathcal{L}_{InfoNCE} = - \frac{1}{T} \sum_{i=1}^{T} \log \frac{\exp\left(\frac{sim(\mathcal{F}^{*} (T_i), \hat{e_j} (T_i)}{\tau}\right)}{\sum_{k=1, i \neq k}^{T} \exp\left(\frac{sim(\mathcal{F}^{*} (T_i), \hat{e_j} (T_k)}{\tau}\right)}
\end{equation}
\vspace{-1cm}


\begin{equation}
\mathcal{L}_{align} = \sum_{i=1}^{5} \mathcal{L}_{InfoNCE}, \quad \small i \in \{\text{P, D, M, O, txt}\}
\end{equation}

Next, we aim to identify and prioritize the most relevant modalities for each snippet by reducing cross-modal conflicts and noise, resulting in task-oriented multi-modal embeddings. First, we \textit{concatenate} the aligned embeddings from each modality along the embedding dimension. Then, we use a stack of \textit{transformer} blocks to highlight the most pertinent modalities by explicitly encoding the cross-correlations among them. Additionally, the RGB embeddings \( F^{*} \) from the student model are added between the transformer blocks to enhance the RGB representation with contextually relevant information from multiple modalities.



Second, we guide the final multi-modal output from the last transformer block, \( \mathcal{F}_{M}^* \in \mathbb{R}^{T \times D_i} \), towards a task-specific representation for WSVAD. This ensures that relevant modalities are produced in the PMG with minimal noise, and that the alignment between salient modalities and RGB is optimized for WSVAD with minimal cross-modal conflict. This is achieved through a distillation process, which minimizes the difference between \( \mathcal{F}_{M}^* \) and the teacher's pre-trained features at the same stage, \( \mathcal{F}_{teach} \in \mathbb{R}^{T \times D_i} \). The distillation loss guiding this minimization is defined as:

\begin{equation}
     \mathcal{L}_{distill} =\frac{1}{D_i} \sum_{k=1}^{D_i}(\mathcal{F}^*_{M_k} - \mathcal{F}_{teach_k})^2.
\end{equation}

\subsection{ $\pi$-VAD Optimization}
$\pi$-VAD is optimized in two steps. In the \textit{first step}, the student model, PMG module, and CMI module are warmed up with the $\mathcal{L}_{PMG}$, $\mathcal{L}_{align}$, and $\mathcal{L}_{distill}$ respectively. This ensures that all the components are correctly initialized before optimizing for the actual task and thereby it avoids possible pitfalls in which one of the modalities overpowers the others independently to what information adds to the RGB embeddings. The loss function for the first step is:

\begin{equation}
     \mathcal{L}_{first} = \mathcal{L}_{PMG} + \mathcal{L}_{align} + \mathcal{L}_{distill}.
    \label{eq:firststage}
\end{equation}

In the \textit{second step}, the model is trained on the WSVAD task with the standard MIL loss function used in UR-DMU~\cite{AAAI23URDMU}. In order to avoid the decoupling of the aligned modalities, the final training objective is:

\begin{equation}
    \mathcal{L}_{second} = \mathcal{L}_{MIL} + \lambda_1 \mathcal{L}_{align} + \lambda_2 \mathcal{L}_{distill} + \mathcal{L}_{PMG}
    \label{eq:secondstage}
\end{equation}

where $\lambda_1$ and $\lambda_2$ are hyper parameters that allows us to balance the impact of the distillation and alignment components on the training process. The $\mathcal{L}_{PMG}$ is not balanced by a factor to ensure that the pseudo-modalities generated remain bounded to the ground-truth modalities throughout the training process.
\section{Experiments}
We conduct extensive experiments on two of the most common and challenging VAD datasets publicly available, UCF-Crime~\cite{cvpr18} and XD-Violence~\cite{eccv2020}. For each dataset we extract the pose, depth, semantic, textual and motion features. The audio features contained in the XD-Violence dataset are considered as an additional modality and aligned via PI with the others. Additionally, we test our approach on the MSAD dataset~\cite{msad2024}, a recently published dataset that contains real-world anomalous videos collected in a more diverse set of scenarios compared to UCF-Crime.
\begin{table}[t]
\small
\parbox{\linewidth}{
\centering
\setlength\tabcolsep{2.5pt} 
\begin{tabular}{ll|cc|cc}
    \toprule
    \textbf{Model} & \textbf{Encoder} & \multicolumn{2}{c|}{\textbf{UCF-Crime}} & \multicolumn{2}{c}{\textbf{XD-Violence}}\\
     & & \textbf{AUC} & \textbf{AUC\textsubscript{A}}  & \textbf{AP} & \textbf{AP\textsubscript{A}} \\
    \midrule
        \multicolumn{6}{c}{\textit{SoTA with multi-modality at inference}}\\
        HL-Net~\cite{eccv2020} &    I3D&      82.44 &      -&   - &    -  \\
        HSN~\cite{HSN}  &  I3D&  85.45 &  -  & -   & -   \\
        MACIL-SD~\cite{yu2022modality} & I3D+audio & - & - & 83.40 & - \\
        UR-DMU & I3D+audio & - & - & 81.77 & - \\
        TPWNG~\cite{CVPR24TPWNG} & CLIP & 87.79 & -& 83.68& - \\
        PEMIL~\cite{CVPR24PEMIL} & I3D+Text& 86.83 & - &88.21&-\\
        VadCLIP~\cite{AAAI2024_vadclip}& CLIP& 88.02 & 70.23 &  84.15& -  \\
        \midrule
        \multicolumn{6}{c}{\textit{SoTA with RGB-only at inference}}\\
        \multirow{2}{*}{ MIL~\cite{cvpr18}}&  C3D & 75.41 & 54.25  &  75.68 & 78.61   \\
        &  I3D &  77.42 &   - & -    &   - \\
        RTFM~\cite{iccv21} &   I3D &      84.30 &   62.96 &  77.81 &   78.57  \\
        CLAV~\cite{CVPR23LAA}&  I3D&  86.10& -   &  - & -\\
        UR-DMU~\cite{AAAI23URDMU}&  I3D& 86.97& 70.81  & 81.66 &  83.94\\
        SSRL~\cite{ECCV2022Scale}&  I3D& 87.43&  - & - & -\\
        MSL~\cite{AAAI22MSL}& V-Swin & 85.30 & -  &  78.28 & -\\
        WSAL~\cite{TIP_WSL}&  I3D& 85.38&  67.38  & - & -\\
        ECU~\cite{CVPR23ECU}& V-Swin &  86.22 & -   &  - & -\\
        MGFN~\cite{AAAI23MGFN}&  V-Swin&  86.67 & -   & -  & - \\
        UMIL~\cite{CVPR23UMIL}& CLIP& 86.75 & 68.68 & - & -   \\
        TSA~\cite{ICIP2023CLIP}& CLIP& 87.58 &  -&  82.17& -  \\
	\midrule

    \bf $\pi$-VAD (Ours) & \bf  I3D& \bf  90.33  & \bf 77.77 & \bf 85.37 & \bf 85.79 \\
    &&\textsuperscript{{\color{red}\scriptsize(+2.75\%)}}&\textsuperscript{{\color{red}\scriptsize(+6.96\%)}}&\textsuperscript{{\color{red}\scriptsize(+3.20\%)}}&\textsuperscript{{\color{red}\scriptsize(+1.85\%)}}\\[-1ex]
    
    \bottomrule
\end{tabular}
\vspace{-0.3cm}
\caption{State-of-the-art comparisons on UCF-Crime and XD-Violence in the WSVAD task. The best results are written in \textbf{bold}.}
\label{tab:testing}
\centering
\begin{tabular}{l|cccc}
    \toprule
    \textbf{Model} & \multicolumn{4}{c}{\textbf{MSAD (NeurIPS'24)}} \\
    & \textbf{AUC} & \textbf{AUC\textsubscript{A}}  & \textbf{AP} & \textbf{AP\textsubscript{A}} \\
     \midrule
    RTFM\cite{iccv21} & 86.65 & - & - & - \\
    MGFN\cite{AAAI23MGFN} & 84.96 & - & - & -   \\
    TEVAD\cite{chen2023tevad} & 86.82 & - & - & -  \\
    \midrule
    UR-DMU\cite{AAAI23URDMU} & 85.02 & -  & - & - \\
    UR-DMU {\color{red}*} & 85.78 & 67.95 & 67.35 & 75.30 \\
    \midrule
    \bf $\pi$-VAD (Ours) & \bf 88.68 & \bf71.25 & \bf71.26 & \bf77.86\\
    \bottomrule
\end{tabular}
\vspace{-0.3cm}
\caption{State-of-the-art comparisons on MSAD. {\color{red}*} indicates our own implementation. The best results are written in \textbf{bold}.}
\label{tab:testing_msad}
}
\vspace{-0.5cm}
\end{table}

We follow the established evaluation protocols~\cite{CVPR23UMIL, AAAI2024_vadclip, cvpr18, eccv2020} to measure the performance of $\pi$-VAD on the datasets. For additional information on the experimental settings, we refer to Section \red{A}
of the appendix. 

\subsection{SoTA comparison and analysis}
\begin{figure}[t]
    \centering
    \includegraphics[width=\linewidth]{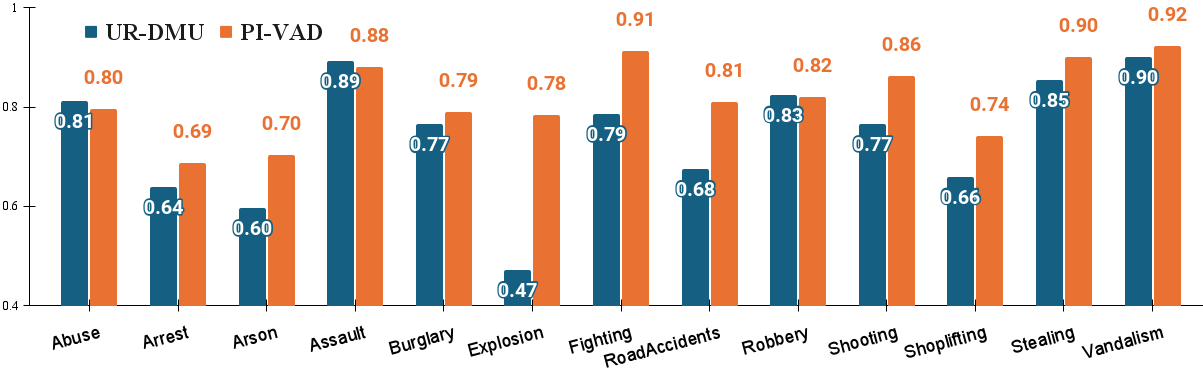}
    \caption{Class-wise $AUC$ comparison of $\pi$-VAD with UR-DMU~\cite{AAAI23URDMU} on the UCF-Crime dataset.}
    \label{fig:classwise}
    \vspace{-0.5cm}
\end{figure}

To evaluate $\pi$-VAD, we place inductor modules at early and late stages of a UR-DMU~\cite{AAAI23URDMU} model. The results of our experiments are shown in Table~\ref{tab:testing}. Compared with current multi-modal SoTA approaches, $\pi$-VAD demonstrates superior capabilities in both UCF-Crime and XD-Violence datasets. On \textbf{UCF-Crime}, $\pi$-VAD marks a $+2.31$\% improvement over multi-modal VadCLIP, and outperforms the best RGB-based model by $+2.75$\%. It is important to notice that the $AUC_A$ metric, which measures the capabilities of the model to detect abnormal events, shows an even larger improvement over the previous methods. In fact,  $\pi$-VAD outscores VadCLIP and UR-DMU, the best scoring multi-modal and RGB-based methods, by $+7.54$\% and $+6.96$\% respectively. In Figure~\ref{fig:classwise} we compare the class-wise performance of  $\pi$-VAD with the baseline model (\textit{i.e.} UR-DMU) employed as a teacher. $\pi$-VAD improves upon the class-wise $AUC$ scores achieved by UR-DMU in all classes except ``Abuse'', ``Assault'' and ``Robbery''. Notably, the ``Explosion'' class proves to be the most challenging class for the UR-DMU, with a score of $47.25$\%. In this class, $\pi$-VAD almost doubles the performance of UR-DMU, which shows the ability of $\pi$-VAD to learn and leverage an extensive poly-modal scene representation towards the detection of short anomaly events. Significant improvements can be observed for other challenging classes, such as ``Shoplifting" and ``Shooting", where the anomalous events happen in subtle ways that are more difficult to be detected by an RGB-only model. 
 
Similar results are observed on the \textbf{XD-Violence} dataset, where $\pi$-VAD improves by $+1.22$\% over the $AP$ score of VadCLIP. The $AP_A$ shows comparable improvements, outperforming UR-DMU by $+1.85$\%. On the more recent \textbf{MSAD} dataset, $\pi$-VAD achieves a $+1.65$\% performance improvement compared to three available SoTA. It is a significant boost on MSAD as it contains 14 diverse scenarios and distinct environmental conditions, presenting a poignant challenge and benchmark for real-world performance. 
We refer to Section \red{B}
of the appendix for a more in-depth analysis of the class-wise performance and contributions from all modalities on these two datasets.


\subsection{Components ablation study}
We analyze the contribution of the two components, PMG and CMI of PI on the overall $AUC$. The core idea behind CMI is that it’s essential to align disparate modalities to form comprehensive scene representation, which can then be adapted for anomaly detection. 
As shown in Table~\ref{tab:loss_ablation} both alignment and adaptation are necessary to fully harness the multi-modal cues. From \textbf{Row-1} it can be observed that when modalities remain decoupled and unguided for VAD, it under perform compared to the baseline model, UR-DMU~\cite{AAAI23URDMU}. Likewise, the model struggles to leverage aligned modalities effectively for the VAD task without a distillation mechanism. This is likely due to residual noise overwhelming salient cues, underscoring the critical role of distillation in filtering and refining the multi-modal cues.
\begin{table}[h]
\small
    \centering
    \begin{tabular}{c|cc|c}
    \toprule
    \textbf{PMG} & \multicolumn{2}{c|}{\textbf{CMI}} & $\mathbf{AUC}$ \\
    & $\mathcal{L}_{align}$ & $\mathcal{L}_{distill}$ & \\
    \midrule
    \checkmark & - & - & 84.66\\
    \checkmark & \checkmark & - & 85.84\\
    \checkmark & - & \checkmark & 86.29 \\
    \midrule
    - & \checkmark & \checkmark & \bf 90.58 \\
    \checkmark & \checkmark & \checkmark & 90.33\\
    \bottomrule
    \end{tabular}
    \vspace{-0.3cm}
    \caption{Contribution of the reconstruction, alignment and distillation auxiliary tasks on the main VAD task for UCF-Crime.}
    \label{tab:loss_ablation}
    \vspace{-0.4cm}
\end{table}


\begin{figure*}[ht]
\begin{center}
\includegraphics[width=1.\textwidth]{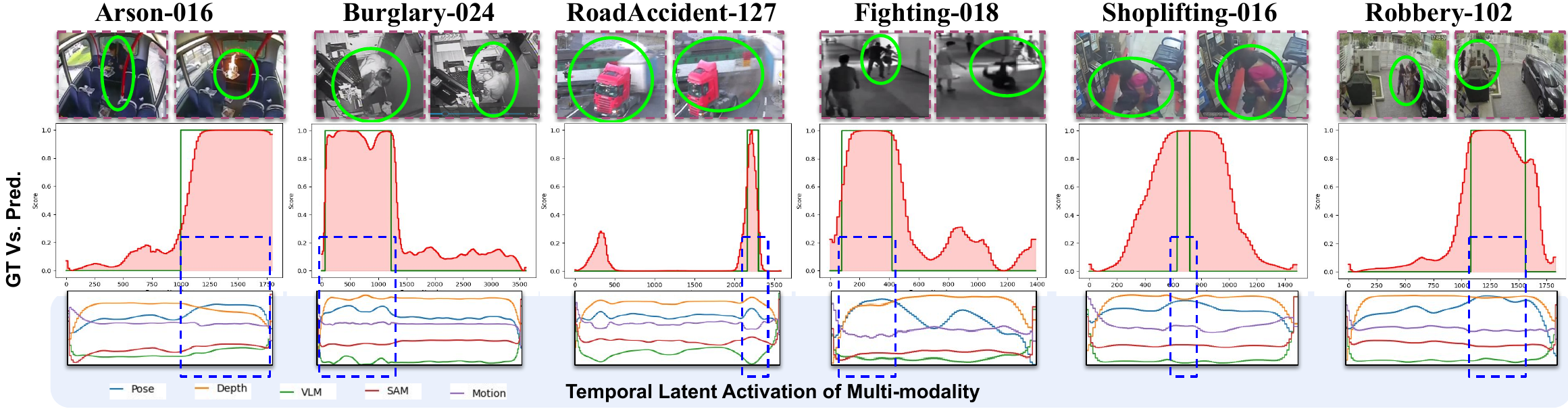}
\vspace{-0.5cm}
\caption{Visualization of sample frames and ground truth (green shed) vs. prediction scores (red shed) for various cases in Row-1 and Row-2. For each plot in Row-2, the X and Y axis denotes the number of frames and corresponding anomaly scores. Row-3 shows the latent activation learned by multi-modality. We plot the mean value of the normalized modalities activations from the first transformer block of the late PI module to show the alignment between modalities and their correlation to the predicted abnormal scores.}
\vspace{-0.7cm}
\label{fig:QA}
\end{center}
\end{figure*}

Reconstructing the modalities at test time leads to a trade-off between computational cost at inference time and performance. Without reconstructing the modalities, PI uses the modality features as input for CMI and obtains a marginal $\simeq0.25\%$ performance gain. The trade-off between computational costs and performances is illustrated in Table~\ref{tab:computation}. Despite requiring more computational and memory resources than the baseline RGB-only method, $\pi$-VAD achieves real-time performance, processing at $30$ frames per second, making it a viable option for practical deployment. 

\begin{table}[h]
\vspace{-0.3cm}
\small
    \centering
    \begin{tabular}{c|ccc}
        \toprule
         & \bf Early & \bf Late & \bf Both \\
        \midrule
       $\mathbf{AUC}$ & 87.14 & 87.48 & \bf 90.33 \\
    \bottomrule
    \end{tabular}
    \vspace{-0.3cm}
    \caption{Comparison of the effect of the early and late PI on performance for UCF-Crime with all modalities available at training.}
    \label{tab:inductors}
    \vspace{-0.4cm}
\end{table}

\begin{table}[h]
\vspace{-0.3cm}
\small
\setlength\tabcolsep{3.5pt} 
    \centering
    \begin{tabular}{c|cc|c}
    \toprule
         & \bf \small UR-DMU & \bf \small Modality Backbones & \small \bf $\pi$-VAD \\
        \midrule
        \bf GFLOPs & 1.54 & 2,561.40 & 19.88\\
        \bf Params. (M) & 6.16 & 2,406.50 & 82.81 \\
        \bf FPS & 110.09 & - & 30.51 \\
        \midrule
        $AUC$ & 86.97 & 90.58 & \bf 90.33 \\
        \bottomrule
    \end{tabular}
    \vspace{-0.3cm}
    \caption{Computational cost comparison with the UR-DMU, the modalities backbones, and the proposed $\pi$-VAD.}
    \label{tab:computation}
        \vspace{-0.5cm}
\end{table}

Further, we observe a complementary effect between early and late PI, as shown in Table~\ref{tab:inductors}. Some anomalies rely on multi-modal association in low-level RGB features, while others need high-level associations. To fully capture all types of anomalies, both early and late PI are essential. Further discussion continues in Section 
\red{C} of the appendix.

\subsection{Qualitative Analysis}
To verify our method's effectiveness, we present qualitative results in Figure~\ref{fig:QA} that illustrate various types of anomalies, including those based on scenes, human actions, and differing durations. Across these scenarios, our method consistently detects anomalies with high confidence, as shown in \textbf{Row-2} of Figure~\ref{fig:QA}. Furthermore, understanding the contribution of each modality to the anomaly scores is essential. \textbf{Row-3} shows two key aspects of multi-modal activation curves: \textbf{(i)} the curve's amplitude and \textbf{(ii)} its pattern in the abnormal regions. The amplitude generally reflects each modality's importance, with depth being critical in CCTV applications. Depth helps distinguish between foreground and background objects, supporting better interaction analysis and occlusion handling.

For scene-based anomalies like ``Burglary-024'' and ``Arson-016'', the pose and text modality activation pattern aligns closely with abnormal regions, due to its strong ability to capture global context. In cases like ``RoadAccident-127'', all modalities contribute significantly to identifying the anomaly. For human-based anomalies, such as ``Fighting-018'' and ``Shoplifting-016'', the pose modality strongly correlates with abnormal regions, and depth complements this by adding spatial context to the 2D key points. Overall, all five modalities are useful for detecting real-world CCTV anomalies, with text and panoptic masks being particularly important for scene-based anomalies, while pose and depth are key for human-based anomalies.



\section{Modality Evaluation}
To analyze the impact of each modality of the VAD task and the interaction between modalities, we focus on the performance of the UCF-Crime. We refer to Section \red{B}
of the appendix for the modality evaluation of the other two datasets.

\subsection{Single Modality Evaluation}
\begin{figure}[t]
    \centering
    \includegraphics[width=\linewidth]{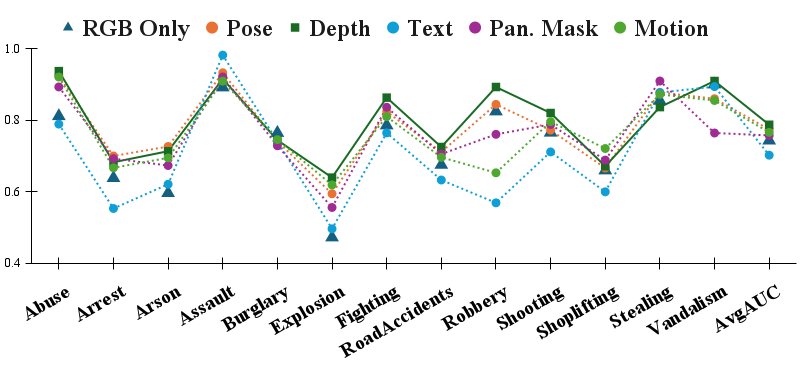} 
    \caption{Class-wise $AUC$ comparison between the RGB  model and RGB with one additional modality model on UCF-Crime.}
    \label{fig:single_classwise}
    \vspace{-0.5cm}
\end{figure}

To properly evaluate the contributions of the different modalities, we trained the model with each modality individually. Table~\ref{tab:modalities_ucf} shows that each modality is able to enhance the RGB features and improve the baseline performance on the UCF-Crime dataset, with motion having the largest positive impact for the $AUC$ metric. This is coherent with the intuitive understanding that motion is often the most important factor in distinguishing between a normal and an abnormal action. However, the depth modality over-performs the others by a large margin on $AUC_A$. The class-wise evaluation for the individual modality contributions in Figure~\ref{fig:single_classwise} shows that depth is the best modality for the majority of classes. We conjecture that the scene information contained in the depth features allows PI to better model the spatial interactions between entities in the scene. This is supported by the performance of the depth modality on the ``Explosion'' class: people and objects tend to move away quickly from the source of an explosion, leading to sharp changes in the depth features.

\begin{table}[t]
\small
\begin{tabular}{ccccc|cc}
    \toprule
    \multicolumn{5}{c}{\textbf{Modality}} & \multicolumn{2}{c}{\textbf{UCF-Crime}} \\
    \textbf{Pose} & \textbf{Depth} & \textbf{Text} & \textbf{Pan.} & \textbf{Motion} & \textbf{AUC} & \textbf{AUC}\textsubscript{\textbf{A}} \\
     \midrule
    - & - & - & - & - & 86.97 & 70.81\\
    \checkmark & - & - & - & - & 87.65 & 73.24 \\
    - & \checkmark & - & - & - & 87.75 & \textbf{75.14} \\
    - & - & \checkmark & - & - & 87.89 & 69.45 \\
    - & - & - & \checkmark & - & 87.71 & 72.13 \\
    - & - & - & - & \checkmark & \textbf{87.92} & 72.04 \\
    \midrule
    \checkmark & \checkmark & - & - & - & 88.14 & 74.06 \\
    \checkmark & \checkmark & \checkmark & - & - & 88.85 & 75.67 \\
    \checkmark & \checkmark & \checkmark & \checkmark & - & 90.31 & 76.47 \\
    \checkmark & \checkmark & \checkmark & \checkmark & \checkmark & \textbf{90.33} & \textbf{77.77}\\
    \bottomrule
\end{tabular}
\vspace{-0.3cm}
\caption{Modality impact comparisons on UCF-Crime. The best results are written in \textbf{bold}.}
\label{tab:modalities_ucf}
\vspace{-0.5cm}
\end{table}

The text modality exhibits robust performance, second-best ranking in the $AUC$ metric and excelling in capturing normal scenarios. We hypothesize that this advantage derives from the ViFiCLIP training, optimized through video-text pairs in the Kinetics-600 dataset~\cite{kinetics}. However, the text modality underperforms in $AUC_A$, likely due to its inclination to represent coarse-grained normalcy patterns more effectively than the details of anomaly events.




\subsection{Poly-modal Evaluation}
\begin{figure}[t]
    \centering
    \includegraphics[width=\linewidth]{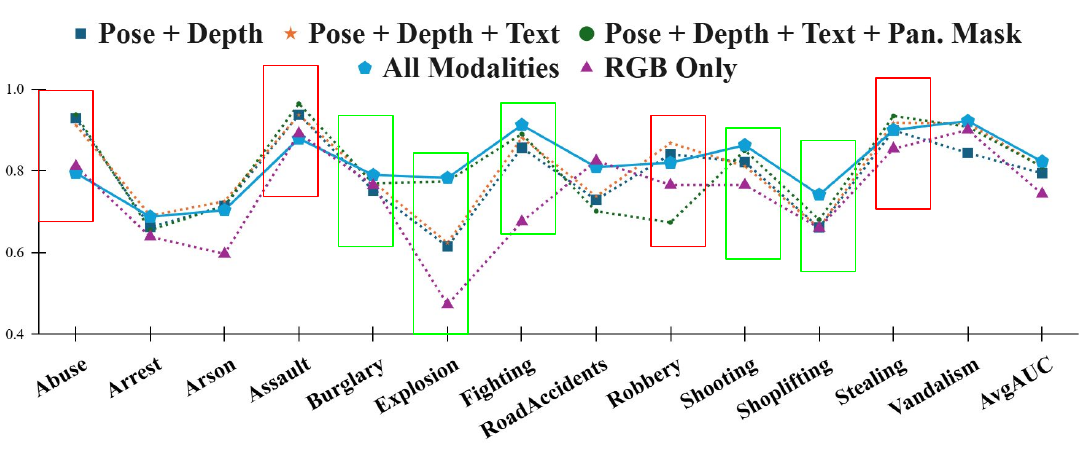}
    \caption{Comparison between the $AUC$ scores of different mixtures of modalities in the $\pi$-VAD framework for the UCF-Crime dataset. In \color{red} red\color{black}, we highlight the classes on which the modalities have contrastive features, in \color{green} green \color{black} the classes where the modalities are complementary.}
    \label{fig:polymodal_classwise}
    \vspace{-0.5cm}
\end{figure}

Table~\ref{tab:modalities_ucf} shows the benefits of combining multiple modalities for WSVAD, where the performance of $\pi$-VAD across the dataset increases by sequentially adding modalities.
Specifically, incorporating textual and panoptic modalities yields the largest performance gains, as they capture essential real-world information from large-scale training datasets, enriching $\pi$-VAD's implicit scene representation. We hypothesize that this representation is comprehensive enough to limit additional contributions from the motion modality for certain types of anomalies, a hypothesis supported by minimal class-wise performance gains in Figure~\ref{fig:polymodal_classwise} and qualitative examples such as “Burglary-024” and ``Fighting-018'' in Figure~\ref{fig:QA}.
The motion modality is nonetheless crucial on other classes, such as ``Shoplifting", where anomalous events are usually subtle and best represented by motion-based cues after the pose cue.

Furthermore, Figure~\ref{fig:polymodal_classwise} suggests that the interaction between different modalities within $\pi$-VAD can lead to complementary or contrastive performance for specific anomaly types. In fact, the modality activations for the ``Robbery-102" example in Figure~\ref{fig:QA} show that pose is the most relevant modality for the anomalous event in the video, while the contributions of the other modalities are marginal. This effect showcases the ability of $\pi$-VAD to leverage the relevant modalities for each anomaly type without over-relying on specific modalities. 



\vspace{-0.3cm}
\section{Conclusion}
\vspace{-0.1cm}
This paper presents $\pi$-VAD, the first poly-modal framework for WSVAD, significantly advancing video anomaly detection by expanding beyond traditional RGB-based methods and integrating multiple modalities to address complex anomaly categories in real-world settings. $\pi$-VAD incorporates five auxiliary modalities, namely pose, depth, panoptic masks, optical flow, and text cues, which jointly enrich anomaly detection with diverse, fine-grained contextual cues. Both novel integrated modules, Pseudo-modality Generator and Cross Modal Induction, enable effective multi-modal learning during training, without imposing additional computational burden during inference. $\pi$-VAD sets a new benchmark for robust and efficient weakly-supervised anomaly detection in real-world applications by showcasing state-of-the-art results on three major datasets.
\vspace{-0.5cm}
\paragraph{Acknowledgements:} This work was supported by Toyota
Motor Europe (TME) and the French government, through
the 3IA Cote d’Azur Investments managed by the National Research Agency (ANR) with the reference number ANR-19-P3IA-0002. We thank Dominick Reilly and Srijan Das for their nice work $\pi$-ViT and inspiration.

{
    \small
    \bibliographystyle{ieeenat_fullname}
    \bibliography{main}
}

\end{document}